\documentclass[letterpaper]{article} % DO NOT CHANGE THIS
\usepackage{aaai2026}  % DO NOT CHANGE THIS
\usepackage{times}  % DO NOT CHANGE THIS
\usepackage{helvet}  % DO NOT CHANGE THIS
\usepackage{courier}  % DO NOT CHANGE THIS
\usepackage[hyphens]{url}  % DO NOT CHANGE THIS
\usepackage{graphicx} % DO NOT CHANGE THIS
\usepackage{natbib}  % DO NOT CHANGE THIS AND DO NOT ADD ANY OPTIONS TO IT
\usepackage{caption} % DO NOT CHANGE THIS AND DO NOT ADD ANY OPTIONS TO IT
\usepackage{algorithm}
\usepackage{algorithmic}
\usepackage{amsmath,amssymb}
\usepackage{subcaption}
\usepackage{booktabs}
\usepackage{color}
\usepackage{amsthm}
\newtheorem{example}{Example}
\newtheorem{theorem}{Theorem}

\newtheorem{assumption}{Assumption}
\newtheorem{lemma}{Lemma}
\newtheorem{corollary}{Corollary}
\newtheorem{proposition}{Proposition}

\usepackage{newfloat}
\usepackage{listings}
\DeclareCaptionStyle{ruled}{labelfont=normalfont,labelsep=colon,strut=off} % DO NOT CHANGE THIS
\floatstyle{ruled}
\newfloat{listing}{tb}{lst}{}
\floatname{listing}{Listing}
\title{Distribution Shift Is Key to Learning Invariant Prediction}

\author{
    Hong Zheng\textsuperscript{1}, Fei Teng\textsuperscript{1,2}\thanks{Corresponding author. \\Author preprint (full version). Accepted at The 40th Annual AAAI Conference on Artificial Intelligence (AAAI 2026).}
}
\affiliations{
    \textsuperscript{\rm 1}School of Computing and Artificial Intelligence, Southwest Jiaotong University, Chengdu, 611756, China\\
    \textsuperscript{\rm 2}Engineering Research Center of Sustainable Urban Intelligent Transportation,\\ Ministry of Education, Chengdu 611756, China \\
   fteng@swjtu.edu.cn
}

\nocopyright  % This version is only for preprint

\begin{document}

\maketitle

\begin{abstract}
An interesting phenomenon arises: Empirical Risk Minimization (ERM) sometimes outperforms methods specifically designed for out-of-distribution tasks. This motivates an investigation into the reasons behind such behavior beyond algorithmic design. In this study, we find that one such reason lies in the distribution shift across training domains. A large degree of distribution shift can lead to better performance even under ERM. Specifically, we derive several theoretical and empirical findings demonstrating that distribution shift plays a crucial role in model learning and benefits learning invariant prediction. Firstly, the proposed upper bounds indicate that the degree of distribution shift directly affects the prediction ability of the learned models. If it is large, the models' ability can increase, approximating invariant prediction models that make stable predictions under arbitrary known or unseen domains; and vice versa. We also prove that, under certain data conditions, ERM solutions can achieve performance comparable to that of invariant prediction models. Secondly, the empirical validation results demonstrated that the predictions of learned models approximate those of Oracle or Optimal models, provided that the degree of distribution shift in the training data increases. 
\end{abstract}

\section{Introduction}
\label{sec1}
Due to distribution shift within the data, Empirical Risk Minimization (ERM) exhibits poor performance~\cite{b9}. To address this, several methods have been proposed for learning invariant prediction in domain generalization (DG), such as Invariant Risk Minimization (IRM)~\cite{b2}, IRM-games~\cite{b7}, among others. The invariant prediction implies stable prediction across environments (domains), which essentially amounts to discovering the labeling mechanism from data~\cite{b1}. Since the causality between covariates $X$ and a response $Y$ reflects the strongest stabilizing associations, discovering such causal relationships appears to be the best choice for learning invariant prediction~\cite{b15}. However, an interesting phenomenon has arisen~\cite{b44}: when carefully implemented, ERM achieves state-of-the-art performance, despite the existence of numerous algorithms specifically designed for out-of-distribution tasks. So, what is the reason causing this? Despite being attributed as a reason for model selection in their work, we argue that there should be a deeper reason behind it. Addressing it could lead to a clearer understanding of the learning process in machine learning. 

Several studies provided explanations for the above phenomenon. One study attributed the phenomenon to the limited number of training domains~\cite{b16}, i.e., poor generalization arises due to insufficient training domains. However, while this explanation accounts for why performance can be poor, it does not explain why ERM can perform well in certain cases with a relatively small number of domains. Since this study made the explanation via their lower bound, we then recall the bounds provided by~\cite{b45}. They provided bounds that consider the complexity of the data condition by using a constant $\lambda$. For a large $\lambda$, it means no classifier performs well on both the testing and training domains, i.e., the data are too complex for generalization; vice versa. Consequently, the good performance of ERM indicates that $\lambda$ is relatively small for certain datasets. However, under DG, we are only allowed to access training domains, indicating that the explanation by $\lambda$ is limited. But it enlightens us that analyzing the training data condition may be a good way. 

In this study, we find that the performance of learning algorithms is driven by distribution shift across training data. A large degree of distribution shift leads to better performance, which is counterintuitive. Through an analysis of a regression case, we observed that the shift in the distributions of the training data (i.e., the joint distribution of $(X, Y)$) determines the learning of regression models. When the shift is large, the learned regression models tend to approach the preset ground-truth model. Inspired by this finding and building on studies~\cite{b41, b40, b1}, we derive several theoretical results, which constitute our contributions as follows: 
\begin{itemize}
\item{ERM solutions exhibit similar performance to invariant prediction models. We prove that the optimal solution of ERM corresponds to an invariant prediction model, provided that the causality-related data assumption holds. 
}
\item{Distribution shift plays a critical role in model learning. The proposed theorems tell us that if one wants to learn predictors from multi-domain data and expects them to perform as well as the realizable or Bayes optimal predictor under each domain, then distribution shift is necessary for such data and should be sufficiently large as the number of data domains increases. 
}
\item{Distribution shift benefits learning invariant prediction. Analytically, a sufficiently large degree of shift among the training domains implies that the domains are well-separated, which in turn ensures that any estimator has minimal generalization error and performs stably, i.e., it approximates invariant prediction models. 
}
\end{itemize} 
According to the above, we can now explain why ERM performs well even with a relatively small number of domains. This is because the introduced data either satisfy a causality-related data assumption or exhibit a large degree of distribution shift in the training domains, making the carefully implemented prerequisite unnecessary under such data conditions. Additionally, we conduct experiments to empirically validate this. The results of a classification task on CMNIST~\cite{b2}, for example, show that the performance of trained models is linearly correlated with the degree of distribution shift, aligning with our main argument. 

\section{Analytical Findings}
\label{sec4}

\subsection{Preliminaries}
\label{sec4.1}
To better understand our work, we first outline several key concepts that will appear in the following sections. 

\noindent\textbf{Distribution shift}: We use the superscript $e$ to denote that a variable is associated with environment $e$, such as $X^e$ and $Y^e$. Following the conventions presented in \cite{b1}, \cite{b2}, and \cite{b3}, we consider the multiple environment data consist of several data environments (domains) indexed by the set ${\cal E}= \{1, 2, \ldots \}$, each associated with several data samples that follow a certain distribution. Correspondingly, the \textit{distribution shift} denotes that ${P^e} \ne {P^{e'}}, \forall e \ne e' \in {\cal E}$. 

\noindent\textbf{Measurement of Distribution shift}: We use the Kullback–Leibler (KL) divergence to measure the difference (i.e., the shift degree) between two given distributions. Other metrics are omitted here, as the KL is only related to our theorem findings. The KL divergence is defined as follows: Let $P$ and $Q$ be two probability measures on a measurable space $(\cal X, \cal A)$. Assume that $P \ll Q$, and denote by $p = dP/d\mu $ and $q = dQ/d\mu$ the Radon-Nikodym derivatives of $P$ and $Q$ with respect to a common dominating measure $\mu$. Then, the KL divergence is defined as 
\begin{align}
KL(P;Q) = \int {\log \frac{p}{q}pd\mu }, \nonumber 
\end{align}
whenever the integral is well-defined. 

\subsection{Deductive Analysis}
\label{sec4.2}
Beginning with a simple case, we present a deductive analysis to explore the factors that determine the performance of ERM solutions. This case aligns with the examples discussed in studies~\cite{b2, b5}, and is as follows: 
\begin{example}
\label{exam1}
Assume that we have several latent variables following Gaussian distributions as follows:
\begin{align}
&{z_1} \leftarrow {\cal N}(0,{a}), {\varepsilon _1} \leftarrow {\cal N}(0,{b}), \nonumber \\
&{z_1} \bot {\varepsilon _1}, {\varepsilon _2} \leftarrow {\cal N}(0,{c}), \nonumber \\
&y = \gamma {z_1} + {\varepsilon _1}, \gamma \leftarrow {\cal N}(0,1), \nonumber \\
&{z_2} = y + {\varepsilon _2}, x = [{z_1},{z_2}]. \nonumber 
\end{align}
We aim to estimate a regression model $\omega$ based on $(x, y)$ by solving $ {\min _{\omega \in {\mathbb R}} }{\mathbb E}[{({\omega ^T}x - y)^2}]$, while the optimal regression model has already been defined above, i.e, ${\omega ^*} = (\gamma \quad {\bf{0}})^T$. According to the normal equation, we have 
\begin{align}
\omega  = \left( {\begin{array}{*{20}{c}}
{\frac{{\gamma a(c + q) - bp - r{p^2} - qp}}{{a(b + c + 2q) - {p^2}}}}\\
{\frac{{1a(b + q)}}{{a(b + c + 2q) - {p^2}}}}
\end{array}} \right) , \nonumber 
\end{align} 
where $p = Cov({z_1},{\varepsilon _2})$ and $q = Cov({\varepsilon _1},{\varepsilon _2})$, and the computational details are provided in the Appendix. Assume $p \to 0$ and $q \to 0$, then we have 
\begin{align}
\label{exam1eq1}
\omega  \approx {\left( {\begin{array}{*{20}{c}}
{\gamma \frac{c}{{b + c}}}&{{\bf{1}}\frac{b}{{b + c}}}
\end{array}} \right)^T}. 
\end{align}   
\end{example}
Based on the result~\eqref{exam1eq1}, we arrive at \textit{a first finding}: the learning of $\omega$ is governed by the distribution $P$ of $x$ and $y$, as $\omega$ is determined solely by the variances $b$ and $c$. Specifically, we find that if $c \gg b$, $\omega \to \omega^*$, as the weight of $\gamma$ approaches $1$ and the weight of $\mathbf{1}$ approaches $0$. In this example, $z_1$, $z_2$, and $\varepsilon_1$ can be regarded as the causal variable, the spurious correlation variable, and the generative noise for $y$, respectively, according to the study~\cite{b15}. For ease of computation, we simply set them to follow Gaussian distributions with zero mean and variances $a$, $b$, and $c$. 

Next, we consider the scenario of multi-domain data, i.e., using $(x^e, y^e)$ for all $e \in \cal E$ to estimate a regression model $\omega$. $\omega^*$ can then be considered as the \textit{invariant prediction model} across domains. Following the settings in Example~\ref{exam1}, i.e., letting $a_e, b_e, c_e$ vary with $e$, we have 
\begin{align}
%\label{exam1eq2}
\omega \approx {\left( {\begin{array}{*{20}{c}}
{{\bf{\gamma }}\frac{1}{{\left| {\cal E} \right|}}\sum\nolimits_{e \in {\cal E}} {\frac{{{c_e}}}{{{b_e} + {c_e}}}} }&{{\bf{1}}\frac{1}{{\left| {\cal E} \right|}}\sum\nolimits_{e \in {\cal E}} {\frac{{{b_e}}}{{{b_e} + {c_e}}}} }
\end{array}} \right)^T} \nonumber 
\end{align} 
Base on the above result, we obtain \textit{a second finding}: the data distributions still determine the learning of regression models, as the weights of $\gamma$ and $\bf{1}$ are essentially similar to the result~\eqref{exam1eq1}, but represent an average over domains. Specifically, if $\sum{c_e} \gg \sum{b_e}$, we have $\omega \to \omega^*$. Based on studies~\cite{b1, b19}, the generative noise $\varepsilon_1$ for $y$ is forbidden to change across all $e \in \cal E$, allowing the causal relationship $z_1 \to y$ to remain invariant. This implies that $\sum{b_e}$ can be treated as a constant. Therefore, we are aware that if the condition $\sum{c_e} \gg \sum{b_e}$ holds, the variance $c_e$ must increase as $e$ grows. This implies that the $KL(P^e;P^{e'}) \ne 0, \forall e \ne e' \in {\cal E}$, as the KL divergence of Gaussian-type variables depends on their mean and variance, and this value should be large. Then, a factor, \textit{distribution shift}, influences learning, completing the deductive analysis.

\section{Theoretical Findings}
\label{sec5}
The analysis results from the previous section interestingly indicate that \textit{distribution shift facilitates the learning of invariant prediction models}. In this section, we provide a theoretical justification for this argument. For each assumption and theorem, we include remarks to help better understanding.

We begin by recalling an assumption outlined in~\cite{b1} to explain what invariant prediction is, as follows: 
\begin{assumption}[ Invariant Prediction~\cite{b1}]
\label{assu1}
There exists a vector of coefficients $ {\gamma ^ * } = {(\gamma _1^ * , \ldots ,\gamma _p^ * )^t}$ with support $ {S^*}: = \{ k:\gamma _k^ *  \ne 0\} \subseteq \{ 1, \ldots ,p\} $ that satisfies $\forall e \in {\cal E}$: $X^e$ has an arbitrary distribution and 
\begin{align}
{Y^e} = \mu + {\gamma ^ * }{X^e} + {\varepsilon ^e}, \nonumber
\end{align} 
${\varepsilon ^e} \sim {F_\varepsilon }$ and ${\varepsilon ^e} \bot X_{{S^*}}^e$, where $\mu  \in {\mathbb R}$ is an intercept term, ${\varepsilon ^e}$ is random noise with mean zero, finite variance, and the same distribution ${F_\varepsilon }$ across all $e \in {\cal E}$. 
\end{assumption}
\noindent\textbf{Remarks}:  (a) The nature of an invariant prediction model is that of a \textit{labeling function}, ensuring stable predictions across domains. (b) Since the error $\varepsilon^e$ follows Gaussian white noise, estimating $\hat{\gamma}$ is feasible for both log-likelihood and least-squares problems. (c) The variable $ X_{{S^*}}^e = {\gamma ^ * }{X^e}$ differs from the causal variable, as this assumption does not imply that $PA(Y^e) = X_{{S^*}}^e $, where $PA(Y^e)$ denotes the parent of $Y^e$ in a directed acyclic causal graph. Consequently, considering invariant prediction as causal prediction is theoretically problematic. Instead, \cite{b1} suggests that causal prediction is a special case under this assumption. 

Then, we consider the data scenario involving causality between $X^e$ and $Y^e$, which is similar to the data structure in Example~\ref{exam1}, and present the following data assumption. 
\begin{assumption}
\label{assu2}
Assume a collection of datasets from multiple environments indexed by the set $\cal E$, where each environment $e \in \cal E$ is associated with an arbitrary distribution on ${\cal X} \times {\cal Y}$. We then assume that the causality between ${X^e}$ and $Y^e$ remains linearly invariant to changes in $e$ and other latent variables in ${X^e}$.  
\end{assumption} 
\noindent\textbf{Remarks}: (a) The causal relationship between $X^e$ and $Y^e$ is required to remain unaffected not only by $e$ but also by any latent variables. This excludes situations involving confounders, mediators, or other variables that can affect the graph from $PA({Y^e}) \to Y^e$. (b) This assumption is, in fact, limited, as real-world data often surprise us with their complex structures. Meanwhile, the invariance implies that $PA({Y^e}) \subseteq {X^e}, \forall e \in \cal E $ holds. 

Based on this causality-related data assumption, we have the following proposition of interest.  
\begin{proposition}
\label{prop1}
Assume $({X^e},{Y^e}),\forall e \in {\cal E}$, with $ \left| {{\cal E}} \right| \ge 1$, satisfies Assumption~\ref{assu2}. Then, the optimal solution $ {\omega ^ * }$ of the learning objective 
\begin{align}
\mathop {\min }\limits_{\omega  \in {{\mathbb R}^n}} \sum\nolimits_{e \in {\cal E}} {\left\| {{\omega ^T}{X^e} - {Y^e}} \right\|} _2^2 \nonumber
\end{align}
satisfies Assumption~\ref{assu1}, namely $ {\gamma ^ * } = {\omega ^ * }$. 
\end{proposition} 
\noindent\textbf{Remarks}: (a) This proposition reveals that ERM solutions can also be considered invariant prediction models, such that they perform well in DG, provided that the data conditions satisfy Assumption~\ref{assu2}. (b) There are several examples of such data, including those used for simple causal analyses in medical diagnosis, air quality assessment, etc. 

The proof of this proposition, along with the proofs of the subsequent theorems, is provided in the Appendix. 
The above proposition, however, does not take into account the main argument concerning data distributions. To address this, we then consider the following assumptions for noise-free and Massart-noisy distribution scenarios. 
\begin{assumption}[Clean Distributions]
\label{assu3}
Assume a hypothesis space $ {\cal H} \subseteq \{ h:{\cal X} \to {\cal Y}\} $, and let $ {\cal P} = \left\{ {P \in {\cal P}({\cal H})} \right\}$ denote a family data distributions, where $ {\cal P}({\cal H}) \subseteq {\cal P}({\cal X} \times {\cal Y})$. 
\end{assumption} 
\noindent\textbf{Remarks}: (a) This assumption implies that the distribution family $\cal P$ is induced by certain functions in $\cal H$, thereby transforming the problem of probability estimation into a function estimation problem. Correspondingly, given the realizable functions, the associated distributions can be determined. (b) Note that for the quantity of data samples in any assumed distribution, we consider it is sufficient and do not mention it thereafter. 

\begin{assumption}[Massart-noisy Distributions]
\label{assu4}
Assume a hypothesis space $ {\cal H} \subseteq \{ h:{\cal X} \to \{ 0,1\} \} $. For $m \in [0,1]$, let ${{\cal P}^m} = \left\{ {P \in {\cal P}({\cal H}):\left| {2\eta (x) - 1} \right| \ge m{\text{ for all }}x \in {\cal X}} \right\}$ denote the family of data distributions satisfying the Massart noise condition~\cite{b40}, where $ {\cal P}({\cal H}) \subseteq {\cal P}({\cal X} \times \{ 0,1\} )$, and $\eta (x) = {\mathbb E}[Y|X = x] = P(Y = 1|X = x)$. 
\end{assumption} 
\noindent\textbf{Remarks}: (a) This assumption aligns with the setting considered in prior studies~\cite{b40, b16} for binary classification. (b) Under this assumption, the Bayes-optimal classifiers remain valid for all distributions in ${\cal P}^m$~\cite{b16}. (c) The Massart noise condition constrains labeling randomness by requiring that, for any $x \in {\cal X}$, the label distribution must not be perfectly ambiguous. Specifically, when $m = 0$, the condition imposes no constraint, permitting arbitrary noise; when $m = 1$, it corresponds to a noise-free scenario. Hence, the Massart noise regime interpolates between the realizable and agnostic settings. 

Assumption~\ref{assu4} is specifically related to the binary classification problem, whereas we do not mention any problem type in Assumption~\ref{assu3}. To measure the difference between any $h \in \cal H$, we present the following measurement assumption. 
\begin{assumption}
\label{assu5}
Let $\cal H$ be a hypothesis space consisting of measurable functions $h:{\cal X} \to {\cal Y}$, where $\cal X$ and $\cal Y$ are measurable spaces, and $\mu$ be a probability measure on $\cal X$. 
Define a distance between any two hypotheses $h, h' \in {\cal H}$ by the ${{\mathbb L}_1}(\mu )$-norm:
\begin{align} 
\label{assu5eq1}
d(h,h') := {\left\| {h(X) - h'(X)} \right\|_1} .
\end{align}
We assume that this distance is uniformly bounded, i.e.,
\begin{align}
0 \le d(h,h') \le \beta, \quad \forall h, h' \in {\cal H}, \nonumber 
\end{align}
for some finite constant $\beta > 0$.  
\end{assumption}
\noindent\textbf{Remarks}: (a) By definition of the distance, $d(h,h') = 0$ if and only if $h(X) = h'(X)$ $\mu$-almost everywhere on $\cal X$. (b) Note that we assume the distance is upper bounded only to exclude the case of infinity. 

Based on the above assumptions, we prove the following two theorems. 
\begin{theorem}
\label{theo1}
Let ${{\cal P}_E} = \{ {P_1},{P_2}, \ldots ,{P_E}\}  \subseteq {\cal P}$, where $\cal P$ is defined in Assumption~\ref{assu3}. Assume that 
\begin{align}
\mathop {\sup }\limits_{P,P' \in {{\cal P}_E}} \left[ {KL(P;P')} \right] \le \alpha , \nonumber
\end{align}
where $\alpha > 0$ is a finite constant. 
For each $P \in {{\cal P}_E }$, let $s^* \in {\cal H}$ denote a labeling (realizable) function such that $s^*(X) = Y$ almost surely under $P$. Then, we define ${S_E} = \{ s_i^ * |i \in \{ 1, \ldots ,E\} \}$ as the set of such functions. 

Given a metric $d$ satisfying the distance definition~\eqref{assu5eq1} in Assumption~\ref{assu5}, for any estimate $\hat s \in {\cal H}$ under ${{\cal P}_E}$, one has   
\begin{align}
\label{theo1eq1}
\frac{1}{E}\sum\limits_{s \in {S_E}} {{{\bf{1}}_{\left\{ {d(\hat s,s) \ge \varepsilon } \right\}}}}  \le (1 - \sigma ) + \sqrt {\frac{{\log (1/\delta )}}{{2E}}} 
\end{align}
with probability at least $1 - \delta$, where 
\begin{align}
\sigma  = \frac{{\alpha  + \log 2}}{{\log (E - 1)}} \nonumber
\end{align}
and $\delta  \in (0,1)$, if 
\begin{align}
\label{theo1eq2}
\alpha  \le \log \left( {\frac{{E - 1}}{2}} \right)  
\end{align}
for any small constant $ \varepsilon \ge 0$. 
\end{theorem}
\noindent\textbf{Remarks}. (a) The assumption on distributions implies that ${\cal P}_E$ constitutes a relatively compact family of distributions and is KL-bounded. (b) The differences between the assumed realizable functions are not important due to the clean data distribution assumption. (c) Inequality~\eqref{theo1eq1} is governed by $\alpha$ and $E$, indicating that the probability of the prediction error being larger than $\varepsilon$ is bounded by the RHS. (d) Vacuous bound: When $E \to \infty$ and $\alpha = 0$, LHS $\le 1.0$ always holds. Worst-case: When $\alpha = 0$ and $E = 3$, LHS $\le C$, where $C$ denotes a large value for the second term on the RHS. According to these results, the significance of $\alpha$ is emphasized, as when its value is zero there is a risk of obtaining both a vacuous bound and the worst case. (e) Tightness: Inequality~\eqref{theo1eq1}, a Hoeffding-type bound, is tight and converges at the rate of ${\cal O}(1/\sqrt{E})$, provided that $\alpha > 0$. Here, $\alpha > 0$ serves as a prerequisite condition for convergence, and when $\alpha \to \log$ and $E \to \infty$, the LHS $\to 0$. This implies that $\alpha$ should be as large as possible within the given bound. 

\begin{theorem}
\label{theo2}
Let ${\cal P}_E^m = \{ {P_1},{P_2}, \ldots ,{P_E}\}  \subseteq {{\cal P}^m}$, where ${\cal P}^m $ is defined in Assumption~\ref{assu4}, such that 
\begin{align}
\mathop {\sup }\limits_{P,P' \in {\cal P}_E^m} \left[ {KL(P;P')} \right] \le \frac{{2\beta {m^2}}}{{1 - {m^2}}}, \nonumber
\end{align}
$m \in [0,1)$, under Assumption~\ref{assu5}. 
Let $H_E$ denote the set of Bayes optimal classifiers corresponding to distributions in ${\cal P}_E^m $, i.e., 
\begin{align}
{H_E} = \{ h_P^ *  = {{\bf{1}}_{\eta (X) \ge 1/2}}:P \in {\cal P}_E^m\}.  \nonumber
\end{align}
Given a metric $d$ satisfying the distance definition~\eqref{assu5eq1} in Assumption~\ref{assu5}, for any estimate $\hat h \in {\cal H}$ under ${{\cal P}_E^m}$, one has 
\begin{align}
\label{theo2eq1}
\frac{1}{E}\sum\limits_{h \in {H_E}} {{{\bf{1}}_{\left\{ {d(\hat h,h) \ge \varepsilon } \right\}}}}  \le (1 - \sigma ) + \sqrt {\frac{{\log (1/\delta )}}{{2E}}}
\end{align}
with probability at least $1 - \delta$, where  
\begin{align}
\sigma  = \frac{{ 2\beta{m^2} + (1 - {m^2})\log 2}}{{(1 - {m^2})\log (E - 1)}}  \nonumber
\end{align}
and $\delta  \in (0,1)$, if 
\begin{align}
\label{theo2eq2}
\beta  \le \frac{{1 - {m^2}}}{{2{m^2}}}\log \left( {\frac{{E - 1}}{2}} \right)  
\end{align}
for any small constant $ \varepsilon \ge 0$ and $m \ne 0$. 
\end{theorem}
\noindent\textbf{Remarks}. (a) Due to the Massart-noisy condition, the differences between Bayes optimal classifiers affect the distribution shifts (see Proof). (b) The conditions~\eqref{theo1eq2} and \eqref{theo2eq2} indicate that $\sigma \le 1$. (c) Others, such as vacuous bound, worst-case, and tightness, are similar to Inequality~\eqref{theo1eq1}. (d) For estimating $\hat s$ or $\hat h$, we do not specify particular learning methods, such as supervised or semi-supervised learning, nor learning algorithms, such as ERM or IRM. As long as our assumptions are satisfied, the above results hold. (e) \textit{The theorems tell us that if one wants to learn predictors from multi-domain data and expects them to perform as well as the realizable or Bayes optimal predictor under each domain, then distribution shift is necessary for such data and should be sufficiently large under the given constraints as the number of data domains increases}.

\noindent\textbf{Limitations of Theorems}: (a) Inequalities~\eqref{theo1eq1} and \eqref{theo2eq1} do not indicate the effect of the number of data samples in each domain on the estimation. (b) The complexity of the hypothesis space, such as the VC-dimension, is not considered. (c) Require $E \ge 2{e^\alpha } + 1$, where the logarithm is taken to the natural base and $\alpha > 0$.

Despite limitations, the theorems above indicate that distribution shift indeed affects the learning of models in a positive way, provided that some conditions are satisfied. Based on these results, we state the following corollary without proof to conclude our theoretical justification. 
\begin{corollary}
\label{coro1}
Let $\hat s$ be an estimator trained on data sampled from ${\cal P}_E$ (or from ${\cal P}_E^m$ in the binary classification setting). If the distribution shift measure $\alpha$ associated with ${\cal P}_E$ satisfies condition~\eqref{theo1eq2} (or the shift measure $\beta$ associated with ${\cal P}_E^m$ satisfies condition~\eqref{theo2eq2}), then $\hat s \approx \gamma ^ *$, provided that $\alpha$ (or $\beta$) is sufficiently large under the given constraints. Here, $\gamma^*$ serves as an invariant prediction model. 
\end{corollary}
\noindent\textbf{Remarks}:  (a) Distribution shift is key to learning invariant prediction. (b) In the linear case, Assumption~\ref{assu1} holds for $\gamma^*$, under the distribution condition ${\cal P}_E$, we have $d(\hat s, {\gamma ^ * }) \le \varepsilon$. Under ${\cal P}_E^m$, the difference between $\gamma ^ *$ and the Bayes classifier $s^*$ is given by ${\left\| {{\gamma ^ * } - {s^ * }} \right\|_1} = KL({P_{{\gamma ^ * }}};{P_{{s^ * }}})/m\log \left( {(1 + m)/(1 - m)} \right) = K$. Then, we have $d(\hat s,{\gamma ^ * }) \le \varepsilon  + K$ (see Lemma 3 in the Appendix). (c) For the nonlinear case, recalling that the invariant prediction models naturally correspond to the labeling model, the above also holds. 

\section{Empirical Validations}
\label{sec6}
The previous section provides theoretical justifications for a key argument: \textit{distribution shift benefits learning invariant prediction}. In this section, we conduct experiments to validate this key argument, not theorems; therefore, the constraint on $E$ is not strictly enforced. Note that, due to the intervenable nature of real-world data, we only experimented with synthetic data. 

\subsection{Settings}
\label{sec6.1}
We first illustrate the data information (two synthetic datasets: one for regression and the other for classification) and then list learning algorithms used for validation.

\noindent\textbf{Synthetic Data for Regression}: Following the data setting in Example~\ref{exam1}, we build the synthetic data. To control the distributions, we let variances $a$ and $c$ be the variance $e$  and set $b = 1.0$ as a constant. Then, by setting different values for $e$, we construct two datasets: $D1$ and $D2$. The $D1$ dataset has three trainable domains, i.e., $e \in {\cal E} = \{ 1.0,2.0,3.0\}$, and the $D2$ dataset has thirty training domains, i.e., $e \in {\cal E} = \{ 1.0,2.0,3.0,\cdots,30.0\}$. In each domain of both datasets, we have ten thousand samples. Comparing the two datasets, we observe that the degree of distribution shift in $D2$ is greater than in $D1$, not only due to the variance settings but also because of the number of domains. The code used to generate them is provided in the Appendix.  

\noindent\textbf{CMNIST for Classification}: The environment setting of CMNIST~\cite{b2} is originally defined as $ {\cal E}_{all} = \{0.1,0.2,0.9\} $, where the decimal value represents the degree of distribution shift (i.e., the degree of correlation between colors and estimated labels), following the Bernoulli distribution. For instance, if $e=0.0$, all color interventions are entirely correlated with the labels. Conversely, if $e=1.0$, the color interventions are completely independent of the labels. In our experiments, we compare two datasets: $C1$ and $C2$. The dataset $C1$ has training domains ${\cal E} = \{ 0.1,0.5\}$, while $C2$ has ${\cal E} = \{ 0.1,0.2\}$. The domain $e = 0.9$ remains unchanged as the unseen domain. As observed, the degree of shift in $C1$ is larger than that in $C2$. 

\noindent\textbf{Learning Algorithms}: We incorporate the baseline ERM, VREx~\cite{b14}, and ERM++~\cite{b11}, as well as the invariance-based IRM~\cite{b2} and P-IRM~\cite{b3}, and the non-invariance-based FISH~\cite{b12}, EQRM~\cite{b17}, and RDM~\cite{b13}. The rationale for this choice is that these algorithms are open-source, allowing us to observe the learning outcomes of both invariance-based and non-invariance-based methods. Other details refer to the Appendix.

\subsection{Experiments on Synthetic Data}
\label{sec6.2}
\noindent\textbf{Conclusion}: The learned regression model approaches the true model when the training data exhibits a large degree of distribution shift and includes a large number of domains, supporting our main argument. 

\begin{figure}[t]
\newcommand{\subfigSize}{0.49}
\newcommand{\inSize}{1.0}
\centering
     \begin{subfigure}{\subfigSize\linewidth}
     \centering
     \includegraphics[width=\inSize\linewidth]{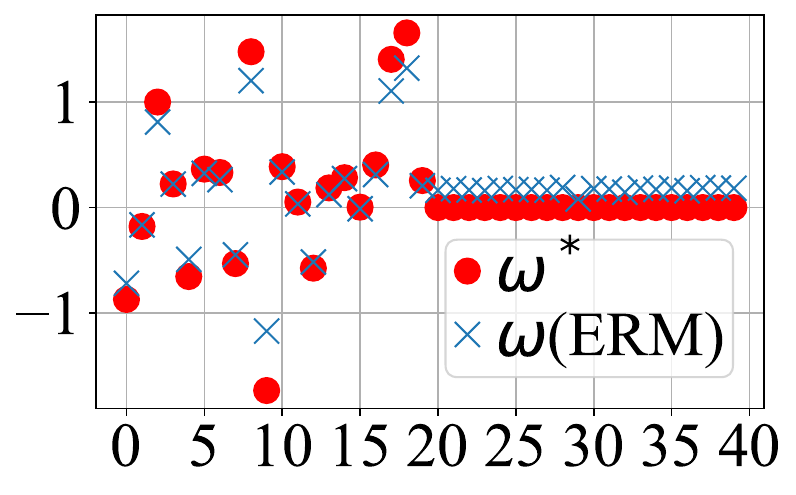}
     \caption{}
     \end{subfigure}
\hfill
     \begin{subfigure}{\subfigSize\linewidth}
     \centering
     \includegraphics[width=\inSize\linewidth]{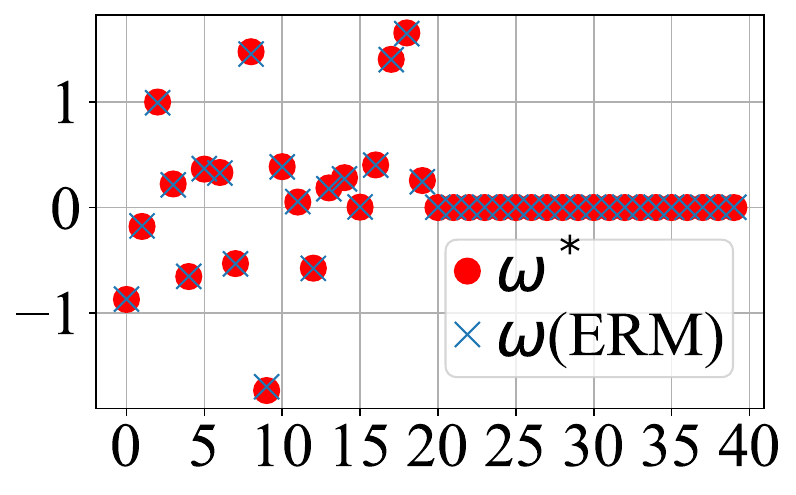}
     \caption{}
     \end{subfigure} 
\caption{Models trained using ERM on the datasets (a) $D1$ and (b) $D2$. As observed, since the shift degree of distributions in $D2$ is larger than that in $D1$, the $\omega$ in (b) approaches $\omega^*$ more closely than the model in (a). Here, $\omega^*$ denotes the preset GT model. }
\label{fig2}
\end{figure}

\noindent\textbf{Experiments}: Since the data is synthetic, we can access the preset GT model. Thus, we observe only the difference between the model $\omega$, trained solely using ERM, and the GT model $\omega^*$, as shown in Figure~\ref{fig2}. Note that $\omega^*$ is an invariant prediction model that satisfies Assumption~\ref{assu1}, as it is the generative model for the response $Y$. As observed from (a) to (b) in this figure, the learned model approximates the true one when the training data has sufficiently large distributional variation. Moreover, the result indicates that increasing the number of domains fundamentally expands the degree of distribution shift. This explains why the trained Machine Learning models perform better when the datasets contain a large number of domains, as presented by study~\cite{b16}. 

\subsection{Experiments on CMNIST}
\label{sec6.2}
\noindent\textbf{Conclusion}: The improvement in evaluation metrics for classification problems benefits from a greater degree of distribution shift in the data, i.e., the performance of trained models is linearly correlated with the degree of distribution shift. This also supports our main argument. 

\begin{table}[t]
\centering
\setlength{\tabcolsep}{1mm}{
\begin{tabular}{lrrrr}
\toprule
{Algorithms} &$e_{0.1}$&$e_{0.5}$&$e_{0.9}$&Mean  \\
\midrule
$ERM$ &\textbf{76.3±2.1}& 66.3±1.6& 49.5±0.9& 64.0±1.5\\
$IRM_{\Omega}$ &58.3±5.6& 55.6±2.7& 50.1±4.9& 54.7±4.4\\
$IRM$ &63.8±5.2&65.0±1.5 & 57.4±4.8&62.1±3.8 \\
$PIRM_{\Omega}$ &59.6±4.6& 53.3±2.8& 44.9±2.8& 52.6±3.4\\
$PIRM$ &65.3±4.3&66.1±2.5 & 58.5±4.2&63.3±3.6 \\
$ERM++$ &\textbf{75.8±1.3}& 67.2±1.6& 49.3±1.7& 64.1±1.5\\
$FISH$ &\textbf{74.6±0.5}& 65.2±1.8& 47.9±0.4& 62.6±0.9\\
$RDM $&70.6±4.1& 66.5±6.1& 54.6±3.5& 63.9±4.6\\
$VREx$ &\textbf{74.7±1.4}& 67.0±1.3& 49.4±2.2& 63.7±1.7\\
$EQRM$ &\textbf{74.3±1.0}& 66.6±1.2& 48.3±0.7& 63.1±1.0\\ 
\hline
\textit{Oracle} & 64.6±1.6 & 66.2±1.4  & 64.5±1.3  & 65.1±1.4  \\ 
\textit{Optimal} & 75.0±0.0 & 75.0±0.0  & 75.0±0.0  & 75.0±0.0  \\ 
\bottomrule
\end{tabular}
}
\caption{Accuracy metrics obtained by different learning algorithms on the testing set for the $C1$ dataset. Here, the values represent the mean and standard deviation. \textit{Oracle} refers to the results obtained by ERM grayscale models, and \textit{Optimal} represents the results for hypothetically optimal invariant models. The \textbf{bold} values indicate abnormal values that exceed the \textit{Optimal} results. }
\label{tab1}
\end{table}

\begin{table}[t]
\centering
\setlength{\tabcolsep}{1mm}{
  	\begin{tabular}{lrrrr}
	\toprule
	{Algorithms} &$e_{0.1}$&$e_{0.2}$&$e_{0.9}$&Mean  \\
	\midrule 
$ERM$ &\textbf{83.6±0.8}& \textbf{79.2±1.2}& 28.3±1.6& 63.7±1.2\\
$IRM_{\Omega}$ &58.2±7.0& 57.0±6.6& 42.8±8.5& 52.7±7.4\\
$IRM$ &\textbf{85.2±1.1}&\textbf{82.1±1.2} &15.2±2.6 &60.8±1.6 \\
$PIRM_{\Omega}$ &64.9±3.8& 64.0±3.9& 32.2±5.0& 53.7±4.2\\
$PIRM$ &\textbf{83.7±3.3}&\textbf{79.7±3.6} &18.4±4.9 &60.6±3.9\\
$ERM++$ &\textbf{84.9±1.4}& \textbf{78.6±1.1}& 27.3±1.3& 63.6±1.3\\
$FISH$ &\textbf{84.6±0.8}& \textbf{79.9±1.3}& 26.3±1.0& 63.6±1.1\\
$RDM$ &\textbf{81.5±2.7}& \textbf{78.2±2.0}& 33.1±6.3& 64.2±4.0\\
$VREx $&\textbf{83.1±1.1}& \textbf{80.8±0.5}& 28.0±2.5& 64.0±1.4\\
$EQRM$ &\textbf{84.1±1.3}& \textbf{78.9±2.5}& 27.8±2.7& 63.6±2.2\\
\hline
\textit{Oracle} & 64.6±1.6 & 66.2±1.4  & 64.5±1.3  & 65.1±1.4  \\ 
\textit{Optimal} & 75.0±0.0 & 75.0±0.0  & 75.0±0.0  & 75.0±0.0  \\ 
	\bottomrule
	\end{tabular}}
\caption{Accuracy metrics obtained by different learning algorithms on the testing set for the $C2$ dataset. }
\label{tab2}
\end{table}

\noindent\textbf{Experiments}: We first validate the classification ability of models trained by different algorithms, which can be seen as factual predictions. Then, we perform hypothesis testing on the factual predictions. Lastly, we validate the classification ability of the models when changing the colors in the original testing data, which can be seen as counterfactual predictions. \textit{Note} that we only use accuracy to evaluate the results, as it sufficiently reflects model performance in both factual and counterfactual prediction scenarios. The Area Under the Curve (AUC) is not informative in this setting, as there is $25\%$ label noise in CMNIST. 

\noindent\textbf{Factual predictions}: The comparison between Tables~\ref{tab1} and \ref{tab2} represents the main result of this validation. As observed, the evaluation metrics on domain $e_{0.9}$ in Table~\ref{tab1} are almost twice as high as those in Table~\ref{tab1}. However, it is NOT important. The \textit{decrease} in evaluations on domain $e_{0.1}$ is the most valuable for this validation experiment. It indicates that the performance of models trained using $C1$ is closer to the oracle model, even the optimal model, under any domain. 

\textit{Notably}, to validate that the improvements observed in the domain $e_{0.9}$ and $e_{0.1}$ are directly related to the distribution changes, we rely on the following hypothesis testing results for validation.

\begin{figure}[t]
\centering
\includegraphics[width=0.8\columnwidth]{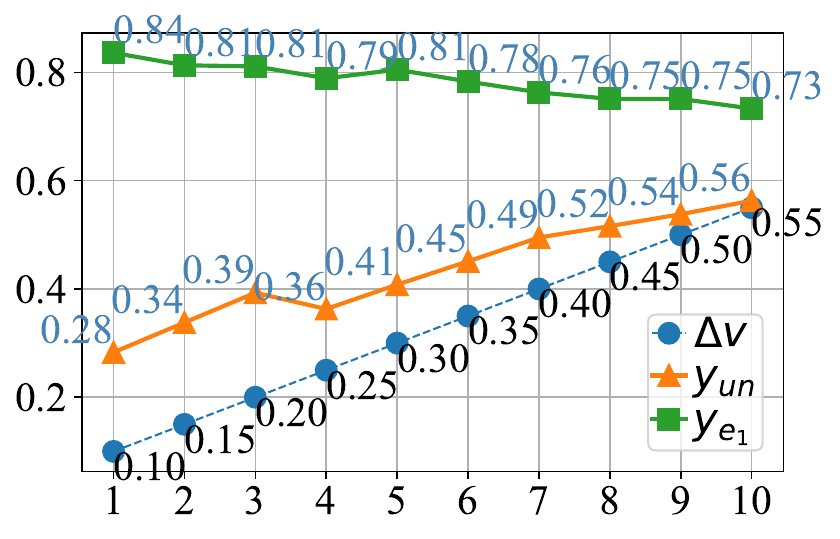} 
\caption{Accuracies ($y_{un}$ and $y_{e1}$) as the shits $\Delta V = \| v_{e_2}-v_{e_1}\|$ increase. All results are obtained from ERM using CMNIST data, where $v_{e_2}$ and $v_{e_1}$ denote the distribution shift settings for spurious correlation colors in domains $e_2$ and $e_1$, respectively. In $10$ sampling trials, we fixed $v_{e_1}$ and varied $v_{e_2}$; thus, the shifts $\Delta V$ increased as sampling progressed. Correspondingly, we obtained $10$ prediction results $y_{un}$ for the unseen domain and $10$ results $y_{e1}$ for domain $e_1$. One observed phenomenon is that as the shifts $\Delta V$ increase, all prediction accuracies tend to converge to values between the oracle and optimal levels. }
\label{fig1}
\end{figure}

\noindent \textbf{Hypothesis Testing for Results on $e_{0.9}$}: First, we state the null hypothesis \\
${H_{{e_{0.9}},0}}$: \textit{The improved results in domain $e_{0.9}$ are unrelated to changes in environmental variables}. \\
The alternative hypothesis aligns with our argument and is omitted here. Second, to validate ${H_{{e_{0.9}},0}}$, we collected data by varying one environmental variable $v_{e_2}$ in a training domain while keeping the $v_{e_1}$ stable, as shown in Figure~\ref{fig1}. The descriptor $v_{e_1}$ represents the values in the first training domain in CMNIST, which are kept unchanged (i.e., $ v_{e_1}=0.1$). The descriptor $v_{e_2}$ represents the values in the second training domain, which are changed (i.e., $ v_{e_2}=\{0.2, 0.25, 0.3, 0.35, \cdots, 0.55 \}$). As observed, the shits $\| v_{e_2}-v_{e_1}\|$ gradually increase. The descriptor $y$ represents the classification accuracy in domain $e_{0.9}$. Then, we can use the parameter $\beta$ to estimate a regression model for these data, specifically $ y = {\beta _0} + {\beta _1}{e_1} + {\beta _2}{e_2} + \varepsilon_{gaussian} $. Using the Python library statsmodels, we obtained the results shown in Table~\ref{tab5}. The p-values are clearly significant, leading us to reject ${H_{{e_{0.9}},0}}$. Additionally, we observed that the t-value of ${\beta _2}$ is larger than ${\beta _1}$, suggesting that $e2$ has a greater impact on the results in the domain $e_{0.9}$.

\begin{table}[t]
\centering
  \centering
  \setlength{\tabcolsep}{3pt}
  	\begin{tabular}{lrrrrrr}
	\toprule
	{} &coef&std err&t&$P>|t|$&[0.025&0.975] \\
	\midrule
${\beta_1}$&1.7856&0.179&9.974&0.000&1.373&2.198\\
${\beta_2}$&0.6029&0.040&15.108&0.000&0.511&0.695\\
	\bottomrule
	\end{tabular}
  \caption{Results of t-testing for ${H_{{e_{0.9}},0}}$.}
  \label{tab5}
\end{table}

\begin{table}[t]
\centering
  \setlength{\tabcolsep}{3pt}
  	\begin{tabular}{lrrrrrr}
	\toprule
	{} &coef&std err&t&$P>|t|$ &[0.025&0.975] \\
	\midrule
${\beta_1}$& 8.7422&0.083&105.669&0.000&8.551&8.933\\
${\beta_2}$&-0.2135&0.018&-11.575& 0.000&-0.256&-0.171\\
	\bottomrule
	\end{tabular}
  \caption{Results of t-testing for ${H_{{e_{0.1}},0}}$.}
  \label{tab6}
\end{table}

\noindent\textbf{Hypothesis Testing for Results on $e_{0.1}$}: Similarly, define the null hypothesis \\
${H_{{e_{0.1}},0}}$: \textit{The decreased accuracy in domain $e_{0.1}$ is unrelated to changes in environmental variables}. \\
The alternative hypothesis, being the opposite, is omitted here. The data for changed and unchanged domains are consistent with the previous hypothesis testing. However, the difference is that we use $y1$ (see Figure~\ref{fig1}) to denote the classification accuracy in domain $e_{0.1}$. We then use the parameter $\beta$ to estimate a regression model for these data, specifically $ y1 = {\beta _0} + {\beta _1}{e_1} + {\beta _2}{e_2} + \varepsilon_{gaussian} $. Using the Python library statsmodels, we obtained the results shown in Table~\ref{tab6}. The p-values are clearly significant, leading us to reject ${H_{{e_{0.1}},0}}$. Moreover, we observed that the t-value of ${\beta_2}$ is negative, indicating that $e2$ is the sole reason for the reduced accuracy in domain $e_{0.1}$. \\
The validation is complete, as supported by the above.

\noindent\textbf{Counterfactual Predictions}: For counterfactual predictions, we altered the colors in the testing data. For example, an original image matrix with a color layer of $[1, 0, 0]$, representing red, was changed to $[0, 1, 0]$, which represents green. This allows us to assess whether the model’s prediction is invariant to non-causal features (color), and hence test its robustness under distributional shift. If predictions vary significantly, this indicates that the model has learned to rely on color as a shortcut, rather than capturing the shape-dependent causal structure. 

\begin{table}[t]
\centering
\setlength{\tabcolsep}{1mm}{
\begin{tabular}{lcrrr}
\toprule
{Algorithms}&{Type}&$e_{0.1}$&$e_{0.5}$&$e_{0.9}$ \\
\midrule
$ERM$&{$CF$}&43.6±2.5&62.6±1.2&76.7±1.8\\
{}&{$F$} &76.3±2.1& 66.3±1.6& 49.5±0.9\\
{}&{$\|CF-F\|$} & \textbf{32.7±0.4}&3.7±0.4& \textbf{27.2±0.9}\\

$ IRM $&{$CF$}&53.9±5.6&65.5±1.4&67.0±4.7\\
{}&{$F$} &63.8±5.2&65.0±1.5 & 57.4±4.8\\
{}&{$\|CF-F\|$} &9.9±0.4&0.5±0.1&9.6±0.1\\

$ PIRM $&{$CF$}&54.7±2.8&66.5±2.0&67.8±3.9\\
{}&{$F$} &65.3±4.3&66.1±2.5 & 58.5±4.2\\
{}&{$\|CF-F\|$} &10.6±1.5&0.4±0.5&9.3±0.3\\

$ RDM $&{$CF$}&50.5±3.0&64.2±7.9&73.9±3.8\\
{}&{$F$} &70.6±4.1& 66.5±6.1& 54.6±3.5\\
{}&{$\|CF-F\|$} &20.1±1.1&2.3±1.8&19.3±0.3\\

\bottomrule
\end{tabular}
}
\caption{Counterfactual (CF) vs. Factual (F) predictions by different methods on the testing set for the $C1$ dataset. The \textbf{bold} values indicate errors exceeding $25\%$.}
\label{tab3}
\end{table}

\begin{table}[t]
\centering
\setlength{\tabcolsep}{1mm}{
\begin{tabular}{lcrrr}
\toprule
{Algorithms}&{Type}&$e_{0.1}$&$e_{0.2}$&$e_{0.9}$ \\
\midrule
$ERM$&{$CF$}&28.9±1.6&36.7±1.8&84.8±0.8\\
{}&{$F$} &83.6±0.8&79.2±1.2& 28.3±1.6\\
{}&{$\|CF-F\|$} & \textbf{54.7±0.8}& \textbf{42.5±0.6}& \textbf{56.5±0.8}\\

$ IRM $&{$CF$}&18.0±2.1&21.5±2.1&87.6±1.4\\
{}&{$F$} &85.2±1.1&82.1±1.2&15.2±2.6\\
{}&{$\|CF-F\|$} &\textbf{67.2±1.0}& \textbf{60.6±0.9}&\textbf{72.4±1.2}\\

$ PIRM $&{$CF$}&20.0±3.6&23.4±4.0&84.7±3.9\\
{}&{$F$} &83.7±3.3&79.7±3.6&18.4±4.9\\
{}&{$\|CF-F\|$} & \textbf{63.7±0.3}& \textbf{56.3±0.4}&\textbf{66.3±1.0}\\

$ RDM $&{$CF$}&33.1±6.5&37.0±4.9&84.9±3.1\\
{}&{$F$} &81.5±2.7&78.2±2.0& 33.1±6.3\\
{}&{$\|CF-F\|$} & \textbf{48.4±3.8}& \textbf{41.2±2.9}& \textbf{51.8±3.2}\\

\bottomrule
\end{tabular}
}
\caption{Counterfactual (CF) vs. Factual (F) predictions by different methods on the testing set for the $C2$ dataset. }
\label{tab4}
\end{table}

The results are exhibited in Tables~\ref{tab3} and \ref{tab4}, which are obtained by four algorithms and are sufficient for illustration purposes. The results of $\|CF – F\|$ exhibit that models trained using $C1$ are less sensible for colors, approach the oracle models, and perform better than those trained using $C2$. We know that the shift degree of distributions in $C1$ is large than $C2$, and the counterfactual prediction results support our main argument. We consider a value of $\|CF – F\|$ greater than $25\%$ to be strongly associated with color, since there is $25\%$ labeling noise in the original test data, and we should allow for up to $25\%$ tolerance under the counterfactual prediction scenario. 

\section{Related Works}
\label{sec2}
We briefly illustrate several studies in Domain Generalization on learning invariant prediction and introduce theoretical bounds that guide generalization. 

\noindent\textbf{Learning Methods}: An invariance-based learning revolution is brewing to address the out-of-distribution problems. For example, there are methods for learning invariant representations~\cite{b6, b8, b10} and causal representations~\cite{b43, b35, b42}. However, several limitations of learning invariance have been discussed by researchers, such as fundamental limits and trade-offs~\cite{b4}, a fundamental design flaw in IRM~\cite{b37}, and fake invariance~\cite{b32}. Moreover, several studies have questioned learning invariance, such as whether the invariances learned by deep neural networks align with human perception~\cite{b34}, and how to learn invariances in the absence of distinct environments~\cite{b38}. Furthermore, some researchers have raised concerns that causal representation learning is inherently ill-posed~\cite{b24}. In conclusion, regardless of the advantages and open questions surrounding learning invariance, it has pushed this field of study a huge step forward. 

\noindent\textbf{Theoretical Bounds}: The theoretical bounds in learning theory provide a guide for how much better the estimator is. From the VC bounds~\cite{b46} to the Massart-noisy bounds~\cite{b40}, the mystery of the learning pattern in machine learning has been gradually revealed. These methods provide several useful fundamental inequalities to analyze different problems, inspiring future studies, including ours. Ben-David et al.~\shortcite{b45} were the first to focus on learning from different domains, providing generalization bounds based on the VC-class hypothesis and the $\cal H$-divergence. However, labeling noise is not considered in this work. A recent study~\cite{b16} addresses this by presenting a lower bound under the VC-class and Massart-noisy condition. All of these works are highly significant for machine learning, as we consider the advancement of fundamental theories to be true progress.

\section{Conclusion}
\label{sec7}
Our findings, in fact, reveal a pessimistic reality: model learning is overly dependent on data conditions. For example, ERM can outperform methods specifically designed for out-of-distribution tasks, provided that certain data conditions are met. When can we develop a learning method that depends less on data conditions? Finally, there are limitations in our work. The real-world data conditions are complex, such that only a few scenarios satisfy Assumption~\ref{assu2}. In most cases, the determining factor is still distribution shift. However, the degree of distribution shift serves as a prior condition, which is typically unknown for most data conditions. 

\section*{Ethical Statement}
This study explores the theoretical mechanisms underlying domain generalization and, incidentally, explains why Empirical Risk Minimization (ERM) sometimes can outperform methods specifically designed for out-of-distribution (OOD) tasks. The insights provided by this study aim to promote a deeper understanding of learning under domain generalization, such as the role of distribution shift in the data, thereby supporting the ethical development and deployment of machine learning technologies for societal benefit. We stress that the outcomes of our research should be utilized with caution, upholding standards of transparency and accountability. 

The positive societal implication of this research lies in understanding how much better or worse the performance of trained machine learning models can be when using complex and variable environmental data, which may originate from autonomous driving, medical diagnosis, financial analysis, or applications in the natural and social sciences. This understanding can promote the development of more robust AI systems for real-world applications. Nonetheless, we acknowledge that improving model robustness across diverse domains may give rise to certain ethical concerns. For example, models with strong generalization capabilities could be deployed in contexts lacking sufficient attention to data privacy, informed consent, or fairness, thereby exacerbating pre-existing social inequalities.

\section*{Acknowledgments}
This research is supported by the grants from the National Natural Science Foundation of China (No.62272398) and Sichuan Science and Technology Program (No.2024NSFJQ0019).

\bibliography{aaai2026}

\onecolumn
\newpage

\appendix
\section{Appendix }

\subsection{Lemmas}

\begin{lemma}[The invariance for Causality~\cite{b1}]
\label{lemm1}
Consider a linear structural equation model for the variables $ ({X_1} = Y,{X_2}, \ldots ,{X_p},{X_{p + 1}})$, with coefficients $ {({\beta _{jk}})_{j,k = 1, \ldots p + 1}}$, whose structure is given by a directed acyclic graph. The independence assumption on the noise variable can be denoted as $ \varepsilon _1^e \bot \{ \varepsilon _j^e;j \in {\bf{{\rm A}{\rm N}}}(1)\} $ for all environment $ e \in {\cal E}$, where $ {\bf{{\rm A}{\rm N}}}(1)$ are the ancestors of $Y$. The Assumption 1 holds for the parents of $Y$, namely $ {S^*} = {\bf{PA}}(1)$, and $ {\gamma ^ * } = {\beta _1}$ , under the assumption: for each $e \in \cal E$, the experimental setting $e$ arises by one or several interventions on variables from $ \{ {X_2}, \ldots ,{X_p},{X_{p + 1}}\} $ but interventions on $Y$ are not allowed.
\end{lemma}

\begin{lemma}[Fano’s lemma~\cite{b41}]
\label{lemm2}
Let $S$ be a finite set of probabilities $S = \{ {P_1};, \ldots ;{P_r}\}$ such that 
\begin{align}
KL({P_i};{P_j}) \le K\quad \forall {P_i},{P_j} \in S. \nonumber
\end{align}
Then for any estimate $\phi $ with values in $\{ 1;2; \ldots ;r\}$ we have 
\begin{align}
\label{lemm2eq1}
{r^{ - 1}}\sum\limits_{i = 1}^r {{P_i}[\phi  \ne i]}  \ge \gamma
\end{align}
if
\begin{align}
\label{lemm2eq2}
K + \log 2 \le (1 - \gamma )\log (r - 1). 
\end{align}
\end{lemma}

\begin{lemma}[\cite{b40}]
\label{lemm3}
Let $h \in [0,1)$, $\mu$ be a probability measure on $\cal X$ and $\cal T$ be a collection of classifiers on $\cal X$. Let $(X, Y)$ be the coordinate mappings on ${\cal X} \times \{ 0,1\}$, and for every $t \in {\cal T}$, define $P_t$ to be the probability distribution on ${\cal X} \times \{ 0,1\}$ such that, under $P_t$, $X$ has distribution $\mu$ and for every $x \in {\cal X}$, $Y$ follows conditionally on $X = x$ a Bernoulli distribution with parameter ${\eta _t}(x)$. Assume that, for some partition ${\cal X} = {{\cal X}_1} \cup {{\cal X}_2}$, one has ${\eta _t}(x) = (1 + (2t(x) - 1)h)/2$ for every $x \in {{\cal X}_1}$ and ${\eta _t}(x)=t(x)=0$ for every $x \in {{\cal X}_2}$. Denoting by $ {\left\|  \cdot  \right\|_1}$ the ${{\mathbb L}_1}(\mu )$-norm, for every $s, t \in {\cal T}$, the Kullback-Leibler information between $P_t$ and $P_s$ is given by 
\begin{align}
KL({P_t};{P_s}) = h\log \left( {\frac{{1 + h}}{{1 - h}}} \right){\left\| {t - s} \right\|_1}.
\end{align}
\end{lemma}

\subsection{Proofs}

\begin{proof}[Proof of Proposition 1]
Without loss of generality, we consider a single domain dataset ${D_e}: = \{ (X_i^e,Y_i^e)\} _{i = 1}^{{n_e}},e \in {{\cal E}}$, where $ {n_e}$ denotes the sample number, namely $X= (X_1^e, \ldots, X_{{n_e}}^e)$ and $Y=(Y_1^e, \ldots, Y_{{n_e}}^e)$. Since $\left| {{{\cal E}_{tr}}} \right| = 1$, we omit the superscript $e$ for clarity in subsequent discussions. Moreover, due to Assumption 2, we have $ PA({Y_i}) \subseteq X_i$, where $PA({Y_i})$ denotes the causal feature for $Y_i$.

We first list the following causal factorization model for $Y$:
\begin{align}
P({Y_1}, \ldots, {Y_n}) = \prod\nolimits_{i = 1}^n {P({Y_i}|P{A}({Y_i}))}. \nonumber
\end{align}
Assume that we use a parameter $\theta$ and a Gaussian white noise $\varepsilon_i$ to model $PA({Y_i})$ from $X_i$, i.e., $PA({Y_i}) = {\theta ^T}{X_i} + \varepsilon_i$. Then, we have 
\begin{align}
\prod\nolimits_{i = 1}^n {P({Y_i}|PA({Y_i}))} &= \prod\nolimits_{i = 1}^n {P({Y_i}|{X_i};\theta )} \nonumber \\
& = \prod\nolimits_{i = 1}^n {\frac{1}{{\sqrt {2\pi } \sigma }}} {e^{ - (\frac{{{Y_i} - {\theta ^T}{X_i}}}{{2{\sigma ^2}}})}}. \nonumber
\end{align}
The second equality holds because $\varepsilon_i$ is Gaussian white noise. However, $\theta$ is unknown, while $X$ and $Y$ are known. Interestingly, the above formula can be considered a likelihood estimation of $\theta$ using $X$ and $Y$, namely
\begin{align}
L(\theta |X,Y) = \prod\nolimits_{i = 1}^n {\frac{1}{{\sqrt {2\pi } \sigma }}} {e^{ - (\frac{{{Y_i} - {\theta ^T}{X_i}}}{{2{\sigma ^2}}})}}. \nonumber
\end{align}

According to this likelihood function, the log-likelihood estimation is given by:
\begin{align}
l(\theta ) = - \frac{n}{2}\ln (2\pi {\sigma ^2}) - \frac{1}{{2{\sigma ^2}}}\sum\limits_{i = 1}^n {{{({Y_i} - {\theta ^T}{X_i})}^2}}. \nonumber
\end{align}
Based on optimization theory, we know that when $\varepsilon_i$ is Gaussian white noise, solving $\max l(\theta)$ is equivalent to solving a least-squares problem, i.e.,
\begin{align}
\mathop {\min }\limits_{\theta \in {{\mathbb R}^n}} \frac{1}{2}\left\| {{\theta ^T}X - Y} \right\|_2^2 \leftrightarrow \mathop {\max }\limits_{\theta \in {{\mathbb R}^n}} l(\theta ) \nonumber
\end{align}
The optimal solution $\theta ^ *$ of the least-squares problem is also the optimal solution to $\max l(\theta)$. 

This is very interesting. We originally aimed to learn a parameter $\theta$ to solve for the causal feature $PA(Y)$ (i.e., the parent of $Y$), but $\theta$ is ultimately used to predict $Y$ using $X$. This indicates that, as long as the data satisfy Assumption 2, learning parameters to predict $Y$ amounts to learning the causal relationship between $X$ and $Y$. Therefore, based on Lemma~\ref{lemm1}, we know that $S^*=PA(Y) = {\theta^{*T}}X + \varepsilon$, and ${\gamma ^ * } = {\theta ^ * }$ in one domain $e$. 
Then, for multiple domain considerations, one can use mathematical induction to complete the proof. Here, we omit the details and give the result 
\begin{align}
{\gamma ^*} = {\theta ^ * } = \mathop {\arg \min }\limits_{\theta  \in {{\mathbb R}^n}} \sum\limits_{e \in {\cal E}} {\left\| {{\theta ^T}{X^e} - {Y^e}} \right\|_2^2}. \nonumber
\end{align}

\end{proof}

\begin{proof}[Proof of Theorem 1]
By Assumption~3, the set ${\cal P}_E$ can be considered as ${{\cal P}_E} = \{ {P_{s_1^ * }},{P_{s_2^ * }}, \ldots ,{P_{s_E^ * }}\} $. Then, under the assumption: ${\sup _{P,P' \in {{\cal P}_E}}}\left[ {KL(P;P')} \right] \le \alpha$, by applying Lemma~\ref{lemm2} and considering that all hypostases are parameterized, for any estimate $\hat s$ with values in ${S_E}(X) = \{ s_i^ *(X) |i \in \{ 1, \ldots ,E\} \}$, we have 
\begin{align}
\frac{1}{E}\sum\limits_{i = 1}^E {P[\hat s \ne {s_i}]}  &\ge \gamma \nonumber 
\end{align}
if $ \alpha  + \log 2 \le (1 - \gamma )\log (E - 1)$. Here, $\hat s \ne {s_i}$ denotes $\hat s(X) \ne {s_i}(X)$. Hereafter, we omit the symbol $*$ from $s_i^*$, as it does not affect the subsequent content. 
Note that the original estimation problem in Lemma~\ref{lemm2} is that given any estimate $ \phi$ and any $X$, we can infer that this $X$ belongs to, i.e., $\phi(X) \to i, P_{\phi(X)} \to P_i, \forall i \in \{1,\cdots,r\}$. Under our problem, similarly, given any estimate $ \hat s$ and any $X$, we can infer that this $X$ belongs to, i.e., $\hat s(X) \to {s_i}(X), \forall i \in \{1,\cdots,E\}$. This process is essentially the same. 
Under Assumption 5, the measurability assumption, we relax the strict equality condition above by allowing a small tolerance $\varepsilon \ge 0$, as follows:
\begin{align}
\frac{1}{E}\sum\limits_{i = 1}^E &{P[d(\hat s,{s_i}) \ge \varepsilon ]}  \ge \gamma,  \nonumber \\
\gamma  \le & 1 - \frac{{ \alpha  + \log 2}}{{\log (E - 1)}}. \nonumber
\end{align} 
Let $\sigma  = (\alpha  + \log 2)/ \log (E - 1) $, and we have 
\begin{align}
\frac{1}{E}\sum\limits_{i = 1}^E {P[d(\hat s,{s_i}) \ge \varepsilon ]}  \ge 1 - \sigma. \nonumber 
\end{align}
This holds by setting ${\gamma ^ * } = 1 - \sigma$. Since $\forall \gamma  \le {\gamma ^ * }$ and $LHS \ge \gamma$, then $LHS \ge \gamma^*$. 

Let $ {Z_i}: = {{\bf{1}}_{\left\{ {d(\hat s,{s_i}) \ge \varepsilon } \right\}}}$, and we have 
\begin{align}
\bar Z &= \frac{1}{E}\sum\limits_{i = 1}^E {{{\bf{1}}_{\left\{ {d(\hat s,{s_i}) \ge \varepsilon } \right\}}}}, \nonumber \\
{\mathbb E}\left[ Z \right] &= \frac{1}{E}\sum\limits_{i = 1}^E {P[d(\hat s,{s_i}) \ge \varepsilon ]}  \ge 1 - \sigma \nonumber
\end{align}
According to Hoeffding’s inequality: $ P[\bar Z - {\mathbb E}\left[ Z \right] \ge {\epsilon}] \le \exp \left( { - 2E{{\epsilon}^2}} \right)$, we have 
\begin{align}
P[\frac{1}{E}\sum\limits_{i = 1}^E {{{\bf{1}}_{\left\{ {d(\hat s,{s_i}) \ge \varepsilon } \right\}}}}  \ge (1 - \sigma ) + {\epsilon}] \le \exp \left( { - 2E{{\epsilon}^2}} \right), \nonumber 
\end{align}
provided that 
\begin{align}
\alpha  \le \log \left( {\frac{{E - 1}}{2}} \right). \nonumber 
\end{align}
This completes the proof. Note that considering $n$ i.i.d. observations for each $P \in {\cal P}_E$ is unnecessary due to Assumption~3. 
\end{proof}

\begin{proof}[Proof of Theorem 2]
Based on Assumption 4 and applying Lemma~\ref{lemm3}, we have 
\begin{align}
KL({P_t};{P_s}) = m\log \left( {\frac{{1 + m}}{{1 - m}}} \right){\left\| {t - s} \right\|_1} \nonumber
\end{align}
for any ${P_t},{P_s} \in {\cal P}_E^m$, where $t$ and $s$ can be regarded as representing Bayes optimal classifiers in $ {H_E} = \{ h_P^ *  = {{\bf{1}}_{\eta (X) \ge 1/2}}:P \in {\cal P}_E^m\} $. 
Then, under Assumption 5, we have  
\begin{align}
KL({P_t};{P_s}) \le \beta m\log \left( {\frac{{1 + m}}{{1 - m}}} \right). \nonumber
\end{align}
Since the above inequality holds for any ${P_t},{P_s} \in {\cal P}_E^m$, let 
\begin{align}
\mathop {\sup }\limits_{P,P' \in {\cal P}_E^m} \left[ {KL(P;P')} \right] \le \beta m\log \left( {\frac{{1 + m}}{{1 - m}}} \right). \nonumber
\end{align}
Furthermore, due to
\begin{align}
m \le {\tanh ^{ - 1}}(m) \le \frac{m}{{1 - {m^2}}}, \nonumber
\end{align}
we have 
\begin{align}
\mathop {\sup }\limits_{P,P' \in {\cal P}_E^m} \left[ {KL(P;P')} \right] \le \beta m\log \left( {\frac{{1 + m}}{{1 - m}}} \right) \le \frac{{2\beta {m^2}}}{{1 - {m^2}}}, \nonumber
\end{align}
which holds whenever $0 \le m < 1$. 
Then, by applying Lemma~\ref{lemm2}, for any estimate $\hat h$, we have 
\begin{align}
\frac{1}{E}\sum\limits_{i = 1}^E {P[d(\hat h,{h_i^*}) \ge \varepsilon ]}  \ge 1 - \sigma, \nonumber
\end{align}
where  
\begin{align}
\sigma  = \frac{{\beta 2{m^2} + (1 - {m^2})\log 2}}{{(1 - {m^2})\log (E - 1)}}. \nonumber 
\end{align}
provided that 
\begin{align}
\beta  \le \frac{{1 - {m^2}}}{{2{m^2}}}\log \left( {\frac{{E - 1}}{2}} \right),\nonumber 
\end{align} 
$m \ne 0$. The subsequent step follows the proof of Theorem 1 by applying Hoeffding’s inequality, thereby completing the proof. 

\end{proof}

\begin{proof}[Proof of Lemma~\ref{lemm1}]
See the proof of Proposition 1 (p. 7) in the reference~\cite{b1}.
\end{proof}

\begin{proof}[Proof of Lemma~\ref{lemm2}]
See the proof of Fano’s lemma (p. 279) in the reference~\cite{b41}.
\end{proof}

\begin{proof}[Proof of Lemma~\ref{lemm3}]
See the proof of Lemma 7 (p. 2350) in the reference~\cite{b40}.
\end{proof}

\begin{proof}[Computational details for Example 1]

Based on the normal equation and the data setting of Example 1, we have $\omega = {({{\mathbb E}_x}[x{x^T}])^{ - 1}}{{\mathbb E}_{x,y}}[xy]$, ${\mathbb E}[x] = 0$, and ${\mathbb E}[y] = 0$. 

Then, we compute ${\Sigma _x} = {{\mathbb E}_x}[x{x^T}]$ and $Cov(x,y) = {{\mathbb E}_{x,y}}[xy]$ as follows: 
\begin{align}
{\Sigma _x} = \left( {\begin{array}{*{20}{c}}
{Var({z_1})}&{Cov({z_1},{z_2})}\\
{Cov({z_2},{z_1})}&{Var({z_2})}
\end{array}} \right) = \left( {\begin{array}{*{20}{c}}
a&{\gamma a + p}\\
{\gamma a + p}&{{\gamma ^2}a + b + c + 2\gamma p + 2q}
\end{array}} \right), \nonumber
\end{align}

\begin{align}
Cov(x,y) = \left( {\begin{array}{*{20}{c}}
{Cov({z_1},y)}\\
{Cov({z_2},y)}
\end{array}} \right) = \left( {\begin{array}{*{20}{c}}
{\gamma a}\\
{{\gamma ^2}a + b + \gamma p + q}
\end{array}} \right). \nonumber 
\end{align}
Here, $p = Cov({z_1},{\varepsilon _2})$ and $q = Cov({\varepsilon _1},{\varepsilon _2})$. 
According to the process of computing the inverse of a symmetric matrix, we have
\begin{align}
\Sigma _x^{ - 1} = \frac{1}{\kappa }\left( {\begin{array}{*{20}{c}}
{{\gamma ^2}a + b + c + 2\gamma p + 2q}&{ - (\gamma a + p)}\\
{ - (\gamma a + p)}&a
\end{array}} \right), \nonumber
\end{align}
assuming $\kappa = a(b + c + 2q) - {p^2} \ne 0$. 
Then, based on $\omega = \Sigma _x^{ - 1}Cov(x,y)$, we have 
\begin{align}
\omega = \left( {\begin{array}{*{20}{c}}
{\frac{{\gamma a(c + q) - bp - \gamma{p^2} - qp}}{{a(b + c + 2q) - {p^2}}}}\\
{\frac{{{1}a(b + q)}}{{a(b + c + 2q) - {p^2}}}}
\end{array}} \right). \nonumber
\end{align}
We assume that $p \to 0$ and $q \to 0$, but we are sure of that because we need the sum of the weights for $\gamma$ and $\bf{1}$ to be 1.0. Then, for the multiple-domain version of the data in Example 1, consider solving ${\min _{\omega  \in {\mathbb R}}}\sum\nolimits_{e \in {\cal E}} {{\mathbb E}[{{({\omega ^T}{x^e} - {y^e})}^2}]}$, and the weights of $\gamma$ and $1$ will be similar to the above but in the form of an average over domains. 

\end{proof}

\subsection{Empirical Validations}

\noindent\textbf{Hyperparameters of Learning Algorithms}: We did not conduct experiments on hyperparameters to select the optimal results for all learning methods, as this study is not focused on those experiments. All hyperparameters are sourced from the DomainBed website\footnote{https://github.com/facebookresearch/DomainBed}. 

\noindent\textbf{Experimental Settings on Synthetic Data}: We consistently utilized the same hyperparameters throughout all phases of training across all learning methods. These hyperparameters include 5000 epochs, the Adam optimizer, and a learning rate of 0.001. Notably, each method underwent training at least ten times to guarantee robust and dependable results. However, we only presented the results that achieved the minimum out-of-distribution (OOD) risk across all methods. Since we only train a tensor variable $\omega$ and the training process involves tensor multiplication ($\hat Y = \omega^T X +\varepsilon $), the batch size is not incorporated. 

\noindent\textbf{Experimental Settings on CMNIST}: The hyperparameters include 100 epochs, the Adam optimizer, a learning rate of 0.0001, and a batch size of 32. The classifier $f$ comprises a sequence with 10 elements: $ f: = \{ {c_0}{b_0},{c_1}{b_1},{c_2}{b_2},{d_{ - 3}},{d_{ - 2}},{d_{ - 1}},{d_0}\} $, where $cb$ represents convolutional layers integrated with batch normalization, and $d$ signifies dense layers. 

\noindent\textbf{Experimental Conditions}: The entire set of experiments was performed on a single Nvidia RTX 4090 24 GB GPU, utilizing the Tensorflow-gpu 2.6.0 deep learning library.

\begin{figure*}[h]
\newcommand{\inSize}{0.24}
\centering
\includegraphics[width=\inSize\linewidth]{fig/fig6a.pdf}
\includegraphics[width=\inSize\linewidth]{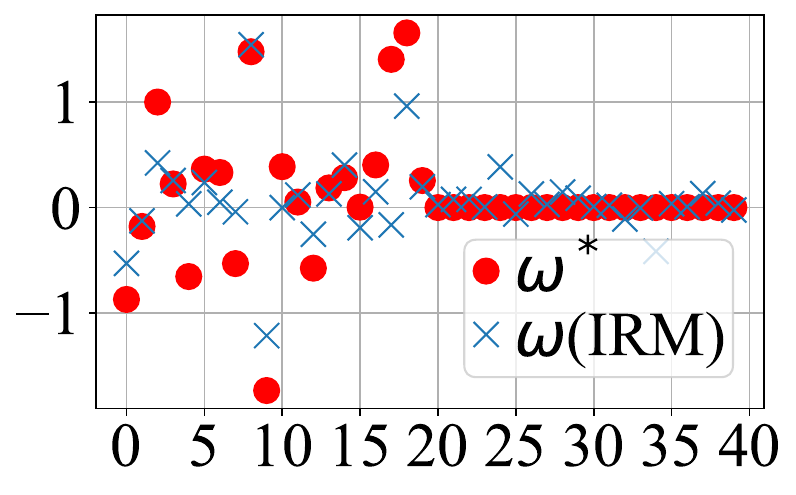}
\includegraphics[width=\inSize\linewidth]{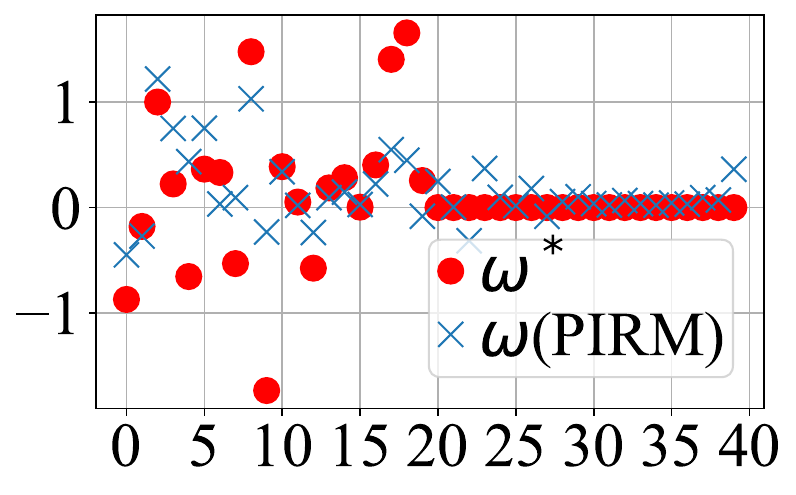}
\includegraphics[width=\inSize\linewidth]{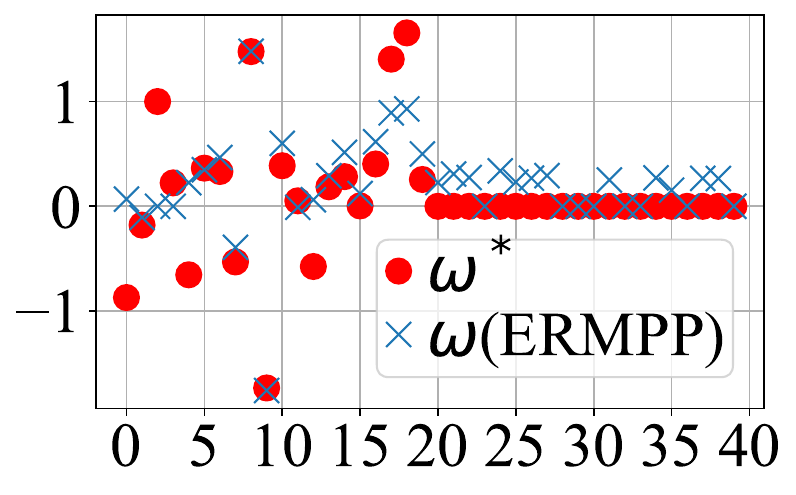}
\includegraphics[width=\inSize\linewidth]{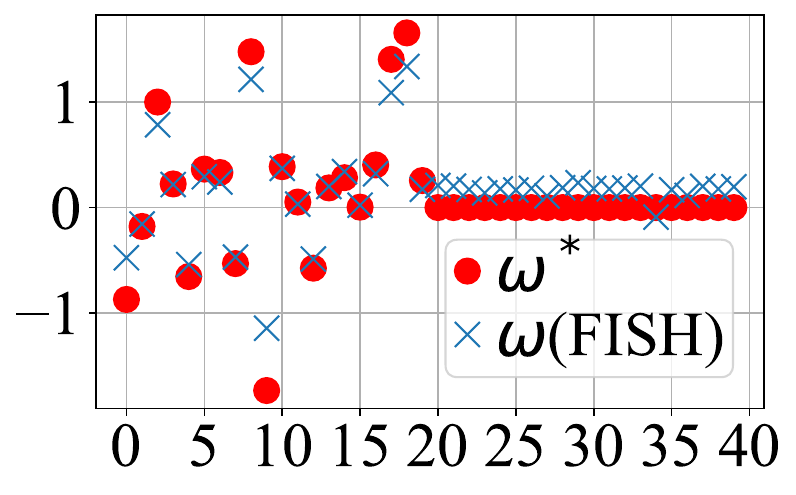}
\includegraphics[width=\inSize\linewidth]{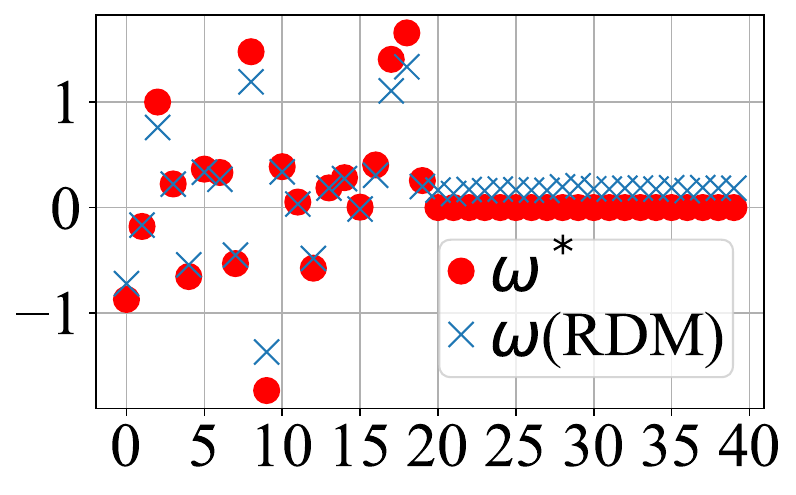}
\includegraphics[width=\inSize\linewidth]{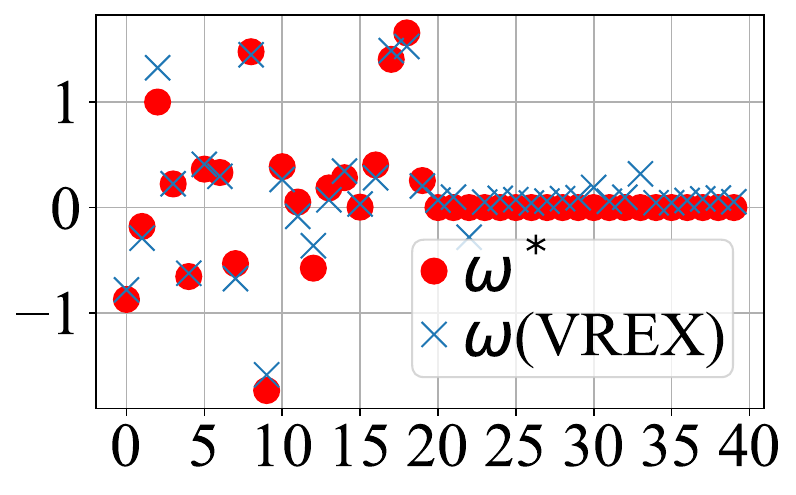}
\includegraphics[width=\inSize\linewidth]{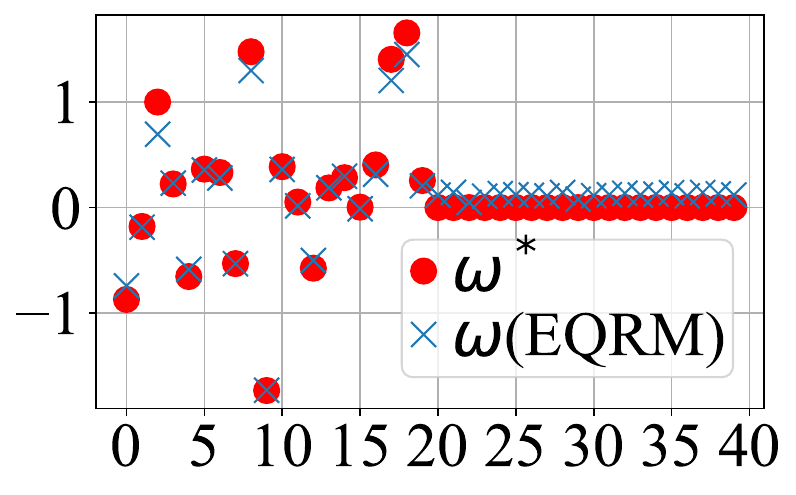}
\caption{Models trained using different learning algorithms on the dataset $D1$.  }
\label{apfig1}
\end{figure*}

\begin{table*}[h]
  \centering
  	\begin{tabular}{lrrrr}
	\toprule
	{Algorithms} &$e_1$&$e_2$&$e_3$&$e_4$ \\
	\midrule
ERM &14.705 / 3.070&16.259 / 3.223&19.448 / 3.523&23.280 / 3.855\\
IRM &23.251 / 3.850&24.407 / 3.940&26.935 / 4.131&29.377 / 4.336\\
PIRM &23.808 / 3.905&25.257 / 4.010&29.099 / 4.317&31.256 / 4.478\\
ERM++ &17.992 / 3.388&20.316 / 3.611&24.582 / 3.951&30.531 / 4.433\\
FISH &14.832 / 3.078&16.480 / 3.249&19.883 / 3.571&23.964 / 3.911\\
RDM &{14.283 / 3.022}&{16.023 / 3.201}&19.303 / 3.518&23.491 / 3.873\\
VREx &17.994 / 3.390&18.628 / 3.452&20.392 / 3.603&22.503 / 3.769\\
EQRM &15.957 / 3.193&16.710 / 3.268&{18.326 / 3.420}&{20.192 / 3.589}\\
	\bottomrule
	\end{tabular}
  \caption{Evaluation metrics obtained by different learning algorithms on the dataset $D1$ (MSE / MAE).}
  \label{aptab1}
\end{table*}

\noindent\textbf{Trained Models on Synthetic Dataset $D1$}: As shown in Figure~\ref{apfig1}, the models trained by different learning algorithms are exhibited. We observe that under the data condition in $D1$, the trained models exhibit distinct characteristics across all methods. The IRM-based models perform better than others in discarding the spurious correlation parts, as the corresponding weights in their results approach $0.0$, whereas the others remain around $0.2$. 

\noindent\textbf{Evaluations on Synthetic Dataset $D1$}: We use only mean squared error (MSE) and mean absolute error (MAE) for evaluation, with the results presented in Table~\ref{aptab1}. As shown, the evaluations for all methods are similar in both known and unseen domains. However, the evaluations of IRM-based methods are less poor than others, indicating the trade-off between learning invariant predictions and minimizing the loss. 

\noindent\textbf{$\Omega$ in IRM-based methods}: IRM-based methods with the subscript $ \Omega $ denote that the scale values in those learning methods are fixed into $1.0$ and are not optimized by gradients. 

\noindent\textbf{Data amounts for the $C1$ and $C2$ datasets }: We only extracted 10\% of data from the original MNIST dataset and divided these data into a training set with 5,996 data samples and a testing set with 996 data samples. The number of data in each \textit{training} domain ($e_{0.1}$, $e_{0.5}$ or $e_{0.2}$) for $C1$ and $C2$ datasets is 2998, while the number of data in each \textit{testing} domain ($e_{0.1}$, $e_{0.5}$ or $e_{0.2}$, $e_{0.9}$) for $C1$ and $C2$ datasets is 332.

\subsection{Codes}
Here, we present only the code for generating the synthetic data in Listing~\ref{lst1}. Other codes, such as training code, algorithm implementations, and Colored-MNIST generation code, are all available online and thus omitted (see the DomainBed website). 

Here, we provide some explanations. The function “Samples” is used to generate the main data based on the introduced ground-truth regression model. Note that the generative noise for $ye$ is set to be orthogonal to $zc$. One can omit this setting and still obtain similar results, but using orthogonality is more general. The function “create\_load\_gamma” is used to create or load a ground-truth regression model; for the first call, please set $rewrite\_gamma=True$. The main function “create\_data” calls the two subfunctions above, and by setting different values in $envs$, one can obtain data following different distributions.   

\begin{listing*}
\caption{Codes for generating synthetic data. }%
\label{lst1}%
\begin{lstlisting}[language=Python]
import os
import numpy as np
import pandas as pd

def Samples(gamma, n=100, d=2, env=1.0):
    zc = np.random.randn(n, d) * env

    noise = np.random.randn(n, d) * env
    dot_product = np.sum(zc * noise)
    noise_orthogonal = noise - dot_product * zc / np.sum(zc * zc)

    ye = gamma * zc + noise_orthogonal

    ze = ye + np.random.randn(n, d) * env

    return np.hstack([zc, ze, np.sum(ye, axis=1, keepdims=True)])

def create_load_gamma(d, path, rewrite_gamma=False):
    if rewrite_gamma:
        print('Rewrite gamma ...')
        gamma = np.random.randn(d)
        df_w = pd.DataFrame(np.hstack([gamma, np.zeros_like(gamma)]))
        name = os.path.join(path, 'true_gamma.csv')
        df_w.to_csv(name, index=False)
        print(name + ' is built!')
    else:
        gamma = pd.read_csv(os.path.join(path, 'true_gamma.csv')).values
        gamma = np.squeeze(gamma[0: d])
    return gamma
    
def create_data(num, d, path, rewrite_data=False, rewrite_gamma=False):
    # generate a new or load the old 

    if not os.path.exists(path):
        os.makedirs(path)

    gamma = create_load_gamma(d, path, rewrite_gamma)
    Data_list = []

    if rewrite_data:
        print('Rewrite data ...')
        envs = [1.0, 2.0, 3.0, 4.0]
        # envs = np.arange(0, 31) + 1.0   # [1.0, ..., 31.0]
        for i, e in enumerate(envs):
            data = Samples(gamma, n=num, d=d, env=e)    
            Data_list.append(np.mean(data, axis=0))

            df = pd.DataFrame(data)
            name = os.path.join(path, f'env{i + 1}.csv')
            df.to_csv(name, index=False)
            print(name + ' is built!')
    return (d * 2)  

if __name__ == "__main__":
    data_root = '/preprocess_data'
    num, d = 10000, 20
    a = create_data(num, d, data_root, rewrite_data=True, rewrite_gamma=False)    
\end{lstlisting}
\end{listing*}

\end{document}